\documentclass[a4paper]{article}

\usepackage{itgspeech2018}    
\usepackage{times}            
\usepackage[english]{babel}   
\usepackage[ansinew]{inputenc}
\usepackage[T1]{fontenc}      
\usepackage[sort&compress,numbers]{natbib}	
\usepackage{amsmath,amssymb}
\usepackage{graphicx}
\usepackage[colorlinks=false,pdfborder={0 0 0}]{hyperref}
\usepackage{units}
\usepackage{xcolor}

\title{Densely Connected Convolutional Networks for Speech Recognition}

\author{Chia Yu Li, Ngoc Thang Vu}

\address{Institute for Natural Language Processing (IMS), University of Stuttgart, Germany\\
  Email: \texttt{\{licu,thangvu\}@ims.uni-stuttgart.de}}

\begin{document}

\maketitle

\begin{abstract}
This paper presents our latest investigation on Densely Connected Convolutional Networks (DenseNets) for acoustic modelling (AM) in automatic speech recognition. DenseN-ets are very deep, compact convolutional neural networks, which have demonstrated incredible improvements over the state-of-the-art results on several data sets in computer vision. Our experimental results show that DenseNet can be used for AM significantly outperforming other neural-based models such as DNNs, CNNs, VGGs. Furthermore, results on Wall Street Journal revealed that with only a half of the training data DenseNet was able to outperform other models trained with the full data set by a large margin.
\end{abstract}


\section{Introduction}
In recent years, deep learning technology has boosted automatic speech recognition (ASR) performance significantly ~\cite{deep-neural-networks-for-acoustic-modeling-in-speech,conversational-speech-transcription-using-context-dependent-deep-neural-networks-2,context-dependent-pre-trained-deep-neural-networks-for-large-vocabulary-speech-recognition}. Various deep learning models have been proposed and developed to improve the performance of ASR systems. Most of them are variations of time delay neural networks (TDNNs)~\cite{waibel1990phoneme}, convolution neural networks (CNNs) ~\cite{abdel2012applying} or recurrent neural networks (RNNs)~\cite{graves2013speech}, or combinations of the two ~\cite{convolutional-networks-for-images-sppech-and-time-series}. Recently, very deep CNNs have demonstrated impressive improvements both in computer vision and speech recognition ~\cite{very-deep-convolutional-networks-for-large-scale-image-recognition,very-deep-convolutional-neural-networks-for-robust-speech-recognition,deep-convolutional-neural-networks-layer-wise-context-expansion-attention}. The main advantage of this architecture is the improved utilization of the computing resources inside the network by a crafted design that allows for increasing the depth and width of the network while keeping the computational budget constant ~\cite{going-deeper-with-convolutions}. Among these, deep CNNs with Layer-wise Context Expansion and Attention architecture (LACE), as presented in ~\cite{deep-convolutional-neural-networks-layer-wise-context-expansion-attention}, are some of the most important variations of CNNs, and consistently outperform other network architectures such as Deep Neural Networks (DNNs), Long Short-Term Memory Networks (LSTMs), bidirectional LSTMs and VG-G networks -  very deep CNN architectures used in image recognition ~\cite{very-deep-convolutional-networks-for-large-scale-image-recognition} - on the 300 hours Switchboard (SWB) Task ~\cite{the-microsoft-2016-conversational-speech-recognition-system}. However, these deep CNNs have a fairly large amount of parameters which need to be trained with a very large amount of training data.

Recently, many papers present complex systems trained on a very large amount of data. For instance, Google uses 18,000 hours of training data for speech recognition for Google Home ~\cite{generation-of-large-scale-simulated-utterances-in-virtual-rooms-to-train-deep-neural-networks-for-far-field-speech-recognition-in-google-home,acoustic-modeling-for-google-home}. There is no doubt that lots of annotated training data can make a system more accurate. However, a massive amount of training data also implies significant computational expense, and makes academic research extremely difficult. Therefore, we aim at finding
compact but deep CNNs that can be trained quickly using limited computational resources. We hope to be able to train accurate ASR systems with such models using a small amount of training data. 

Lately in the computer vision research community, Den-sely Connected Convolutional Networks (DenseNets) have obtained significant improvements over the state-of-the-art networks on four highly competitive object recognition benchmark tasks (CIFAR-10, CIFAR-100, SVHN, and ImageNet) ~\cite{densely-connected-convolutional-networks}. The idea is to introduce shorter connections between layers close to the input and those close to the output which alleviate the vanishing-gradient problem. Furthermore, DenseNet requires fewer parameters than traditional CNNs; because they facilitate feature propagation and reuse features, there is no need to relearn redundant feature maps. For example, a 6 layer DNN with 1024 units for each layer has around 8 Million trainable parameters, while a DensNet with 40 convolutional layers has only around 1 Million parameters to be trained. DenseNets are relatively compact networks and have the structure of deep CNNs, which have been shown to be robust for ASR. Therefore, it is promising to apply DenseNets to acoustic modelling in ASR systems.

This paper is the first one presenting results of DenseN-ets on ASR tasks. Since DensNet is a quite compact network architecture with few trainable parameters, we explore the performance of DenseNet acoustic models trained with Resource Management (RM) and Wall Street Journal (WSJ) data sets and investigate the effect of hyper-parameters for DenseNet models. 
Our experimental results show that the depth of the network plays an important role. 
Besides, the bottleneck and compression components are helpful for speeding up the training. 
A comparison with other network architectures such as DNNs, CNNs and VGGs shows that DenseNet models perform best on RM and WSJ data sets. 
Furthermore, DenseNet achieved impressive performance with only a half of the WSJ training data outperforming all other systems trained with the full data set in our experiment with a large margin.

\section{DenseNet Acoustic Models}
DenseNets obtain significant improvements over the state-of-the-art neural networks on four highly competitive object recognition benchmark tasks (CIFAR-10, CIFAR-100, SVHN, and ImageNet) ~\cite{densely-connected-convolutional-networks}. This is because DenseNets contain shorter connections between layers close to the input and those close to the output. For example, Fig.~\ref{fig:Denseblock} shows that the 3-layer traditional convolutional networks possess 3 connections while 3-layer \textit{dense block} DenseNet have $\dfrac{3(3+1)}{2}$ connections. 
In our experiment, we use DenseNet for acoustic modeling which takes speech features as input and predicts the HMM state which is regarded as sub-phoneme. In this work, we used unadapted features 40-dimensional log Mel filterbank. The important components and hyperparameters for DenseNets are in the following subsections.

\subsection{Dense connectivity}
\begin{figure}[t]
  \centerline{\includegraphics[width=.8\columnwidth]{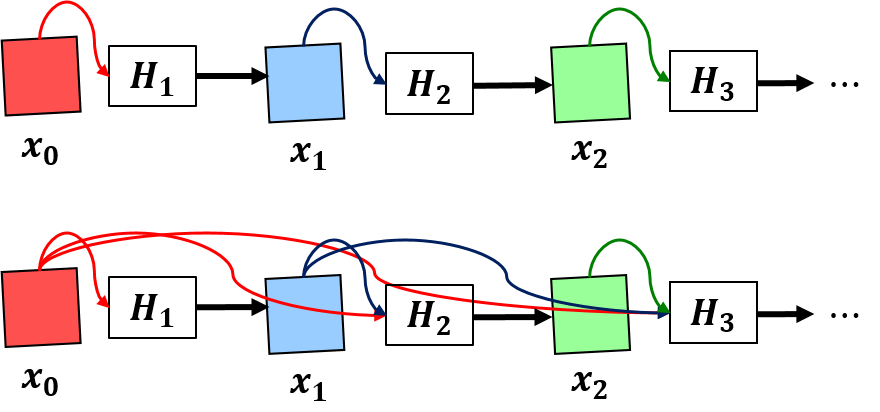}}
  \caption{The examples of 3-layer traditional convolutional networks (top) and 3-layer \text{\it dense block} DenseNet (bottom).}
  \label{fig:Denseblock}
\end{figure}

DenseNet has direct connections from any layer to all subsequent layers. Consider an input feature $x_0$ and a CNN with $N$ layers, where each layer $n$ is equipped with a nonlinear transformation $H_n(\cdot)$. The output of the $n^{th}$ layer is denoted by $x_n$.The output of $n^{th}$ layer is
\begin{equation}
x_n = H_n([x_0,x_1,x_2,...,x_{n-1}]),
\label{Hn}
\end{equation}
where $[x_0,x_1,x_2,...,x_{n-1}]$ refers to the concatenation of the feature maps yielded in layers $0,1,...,n-1$. Fig.~\ref{fig:Denseblock} is the visualization of dense connectivity. Each layer takes all preceding feature-maps as input.

\subsection{Composite function}
The composite function $H_n(\cdot)$ in DenseNet is the composition of three consecutive operations: batch normalization (BN), followed by a ReLU and a $3 \times 3$ convolution.

\subsection{Dense block}
\begin{figure}[t]
  \centerline{\includegraphics[width=1\columnwidth]{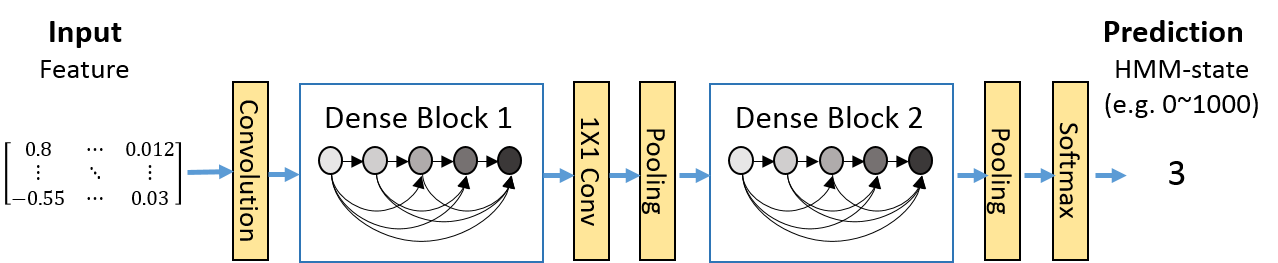}}
  \caption{A deep DenseNet with two dense blocks for Speech Recognition.}
  \label{fig:DenseNet3b}
\end{figure}

We will refer to the structure described above as one layer. Each \text{\it dense block} could contain more than one layer. For example, each \text{\it dense block} in Fig.~\ref{fig:DenseNet3b} contains 5 layers.

\subsection{Transition layer}
A DenseNet consists of multiple \text{\it dense} \text{\it blocks}, connected in series and separated by \textit{transition layers}.
Each transition layer consists of a $1 \times 1$ convolution layer, which is meant to reduce the number of feature maps if the compression is enabled, and a $2 \times 2$ average pooling layer.

The compression is designed to improve model compactness and DenseNet uses one parameter $\theta$, which ranges from 0 to 1, to control the strength of compression. For example, if a \text{\it dense block} has $y$ feature-maps, the transition layer generates $\lfloor \theta y \rfloor$ output feature-maps. Fig.~\ref{fig:DenseNet3b} illustrates how these \text{\it dense blocks} and \text{\it transition layers} are composed in a DenseNet; note that pooling is only performed outside of \text{\it dense blocks}.


\subsection{Growth rate}
The \textit{growth rate} of a DenseNet is the number of channels in its convolution layers. By equation~\eqref{Hn}, the $n^{th}$ layer within a dense block  has $k \times (n-1) + k_0$ input feature-maps, where $k_0$ is the number of input channels and $k$ (the model's growth rate) is the number of channels for subsequent convolution layers. DenseNet has better performance when $k$ is  a small integer, e.g.\ $k = 12$.


\subsection{Bottleneck layer}
In a dense block, a $1 \times 1$ convolution is introduced as bottleneck layer before each $3 \times 3$ convolution to reduce the number of input feature-maps, and thus to improve computational efficiency. In ~\cite{densely-connected-convolutional-networks}, they found this design especially effective for DenseNet. In our experiment, we let each $1 \times 1$ convolution produce $4k$ feature-maps and investigate the effect on Resource Management data set.

In summary, DenseNet is composed of one convolution layer and multiple \text{\it dense block} separated by \text{\it transition layer} shown in Fig.~\ref{fig:DenseNet3b}. The hyperparameters for DenseNets are the kernel size and the number of channels for the initial convolution layer, the number of layers (depth), growth rate, the number of \text{\it dense blocks}. In the experiment, we refer to our model as DenseNet-C or DenseNet-BC when the compression is enabled or both bottleneck and compression are enabled.

\section{Experimental setup}
We conducted several experiments on two data sets (Resource Management(RM) and Wall Stree Journal (WSJ)) to investigate the effect of the Bottleneck and Compression components, the network depth and the number of dense blocks of DenseNets. Furthermore, we aimed at finding the impact of training size on the final performance of DenseNets. 
\subsection{Resources}
Our experiments were performed on different types of corpus: the DARPA 1000-words English language Resource Management(RM) ~\cite{the-darpa-1000-word-resource-management-database-for-continuous-speech-recognition} and the Wall Street Journal (WSJ) ~\cite{the-design-for-the-wall-street-journal-based-csr-corpus}. The complete RM corpus comprises 3 hours of training data (read sentences) and 1
hour of test data. The WSJ data set contains approximately 78 hours of speech which are the results of spontaneous dictation by journalists, and the test sets $dev'93$ and $eval'92$ are 20k-closed conditions. The ASR systems are built up with the Kaldi speech recognition toolkit ~\cite{the-kaldi-speech-recognition-toolkit} and the acoustic models are implemented with PDNN (A Python Deep learning toolkit) ~\cite{kaldi-pdnn},
Theano ~\cite{theano-a-python-framework-for-fast-computation-of-mathematical-expressions}, Lasagne ~\cite{lasagne} and DenseNet source code ~\cite{densely-connected-convolutional-networks}.  The language model used in RM experiments is traditional RM grammar (bigram word-pair), and the one used in WSJ is the 20k vocabulary trigram.

\subsection{Baseline systems}
The baseline systems are Deep Neural Networks (DNN), Convolution Neural Networks (CNN) and very deep Convolution Networks - VGG. DNN contains 6 hidden layers with sigmoid activation functions and 1 fully-connected output layer with a softmax activation. Each hidden layer has 1024 units and is trained using the 40-dimensional log Mel filterbank (FBANK) features. CNN is composed of two convolution layers and max-pooling layers, and 4 affine layers with sigmoid activation function. Each affine layer contains 1024 units and is trained using FBANK with the first and the second time derivatives ~\cite{improving-wideband-speech-recognition-using-mixed-bandwidth-training-data-in-cd-dnn-hmm}. Both DNN and CNN use the same context window (5) and batch size (256). VGG architecture in our experiment is the same as the one used in ~\cite{the-microsoft-2016-conversational-speech-recognition-system}. It contains 14 weight layers which are 3 convolution layers with 96 filters, one max pool, 4 convolution layers with 192 filters, one max pool, 4 convolution layers with 384 filters, one max pool and 2 fully-connected layers at the top. VGG is trained using 40-dimensional FBANK and the context window is 20.

\begin{table}[t]
	\centering
    \scalebox{0.7}{
	\begin{tabular}{l|c|cc|cc}
    \noalign{\hrule height 1pt}

	Layers &  Output& \multicolumn{2}{c}{DenseNet-BC} & \multicolumn{2}{c}{DenseNet/DenseNet-C}\\
		   &  Size  & \multicolumn{2}{c}{(depth=22)}  & \multicolumn{2}{c}{(depth=22)}\\
\hline
\hline
Convolution & $9 \times 38$ & \multicolumn{4}{c}{$3 \times 3$ conv} \\
\hline
Dense block (1)& $9 \times 38$ &  
$\left\{
  \begin{tabular}{c}
    $1 \times 1$ conv\\
    $3 \times 3$ conv
  \end{tabular}\right\}$  
$\times 3$ && 
$\left\{
  \begin{tabular}{c}
    $3 \times 3$ conv
  \end{tabular}\right\}$  
$\times 6$ \\
\hline
Transition (1)& $\begin{tabular}{c} $9 \times 38$ \\ $4 \times 19$ \end{tabular}$ & \multicolumn{4}{c}{
$\begin{tabular}{c}
	$1 \times 1$ conv, $\theta c$ \\ 
	$2 \times 2$ avg. pool
	\end{tabular}$}\\
\hline
Dense block (2)& $4 \times 19$ &  
$\left\{
  \begin{tabular}{c}
    $1 \times 1$ conv\\
    $3 \times 3$ conv
  \end{tabular}\right\}$  
$\times 3$ &&
$\left\{
  \begin{tabular}{c}
    $3 \times 3$ conv
  \end{tabular}\right\}$  
$\times 6$\\
\hline
Transition (2)& $\begin{tabular}{c} $4 \times 19$\\ $2 \times 9$ \end{tabular}$& \multicolumn{4}{c}{
$\begin{tabular}{c}
	$1 \times 1$ conv, $\theta c$ \\ 
	$2 \times 2$ avg. pool
	\end{tabular}$}\\
\hline
Dense block (3)& $2 \times 9$ &  
$\left\{
  \begin{tabular}{c}
    $1 \times 1$ conv\\
    $3 \times 3$ conv
  \end{tabular}\right\}$  
$\times 3$ &&
$\left\{
  \begin{tabular}{c}
    $3 \times 3$ conv
  \end{tabular}\right\}$  
$\times 6$ \\
\hline
Classification & $\begin{tabular}{c}
$1 \times 1$ \\
$$
\end{tabular}$ & \multicolumn{4}{c}{
$\begin{tabular}{c}
	$2 \times 9$ global avg. pool\\ 
	fully-connected, softmax
	\end{tabular}$}\\
\hline
\end{tabular}
}
\caption{The architectures of DenseNet-BC, DenseNet and DenseNet-C with depth=22 and dense blocks=3.}
\label{table:DenseNet_architecture}
\end{table}

\begin{table}[t]
\small
    \centering
	\begin{tabular}{llcccl}
    \noalign{\hrule height 1pt}
	\text{System} & \text{\# of} & \text{depth} & \text{$\theta$} & \text{WER} & \text{Train}   \\
    	 & \text{db.} &	& &			  			& \text{time} 			\\
           &&&								& &\text{\small (hours)}\\
    \hline
	DenseNet	 & 3 & 22 & 1.0 & 2.36 & 8.5 \\
	DenseNet-C   & 3 & 22 & 0.5 & 2.34 & 7.9 \\
    DenseNet-BC  & 3 & 22 & 0.5 & 3.30 & 5\\
    \hline
	DenseNet      & 3 & 41 & 1.0 &  2.23 & 17\\
	DenseNet-C    & 3 & 41 & 0.5 &  \bf{2.12} & 15\\
    DenseNet-BC   & 3 & 41 & 0.5 &  2.51 & 10\\
    \noalign{\hrule height 1pt}
	\end{tabular}
    \caption{Comparison of systems performance with 3 dense blocks and depth on 3 hours Resource Management.}
    \label{table:rm_1}
\end{table}

\begin{table}[t]
\small
    \centering
	\begin{tabular}{llcccl}
    \noalign{\hrule height 1pt}
	\text{System} & \text{\# of} & \text{depth} & \text{$\theta$} & \text{WER} & \text{Train}   \\
    	   & \text{db.} &	& &			  			& \text{time} 			\\
           &&&								& &\text{\small(hours)}\\
    \hline
    DenseNet-C   & 4 & 41 & 0.5 &  2.10  & 2.6 \\
	DenseNet-C   & 4 & 61 & 0.5 &  \textbf{2.05} & 8 \\
    DenseNet-C   & 4 & 81 & 0.5 &  2.14 & 16 \\
	DenseNet-C   & 4 & 101 & 0.5 & 2.45 & 17 \\
	\noalign{\hrule height 1pt}
	\end{tabular}
    \caption{Comparison of systems performance with 4 dense blocks and different depth on 3 hours Resource Management.}
    \label{table:rm_2}
\end{table}

\begin{table}[t]
\small
    \centering
	\begin{tabular}{lccc}
    \noalign{\hrule height 1pt}
	\text{System}  & \text{\# of db., depth} & \text{\# of}  \\
    	           &	  			   & \text{param.} \\
    \hline
	DenseNet	 & 3, 41 & 1.6M &\\
	DenseNet-C   & 3, 41 & 914K &\\
    DenseNet-BC  & 3, 41 & 387K &\\
    DenseNet-C   & 4, 41 & 661K &\\
    \noalign{\hrule height 1pt}
	\end{tabular}
    \caption{An summary of \# of param. for all DenseNet models with depth=41.}
    \label{table:param}
\end{table}

\begin{table}[t]
	\small
    \centering
	\begin{tabular}{llccr}
    \noalign{\hrule height 1pt}
	System & \# of & depth & $\theta$ & WER \\
    	   & db. &	& &			  			\\
           &&&&								\\
	\hline
	DenseNet-C	 & 4 & 61 & 0.1 &  2.13		\\
	DenseNet-C   & 4 & 61 & 0.2 &  2.26		\\
	DenseNet-C   & 4 & 61 & 0.3 &  1.99		\\
	DenseNet-C   & 4 & 61 & \bf{0.4} &  \bf{1.91} 	\\
    DenseNet-C	 & 4 & 61 & 0.5 &  2.05 	\\
	DenseNet-C   & 4 & 61 & 0.6 &  2.05 	\\
	DenseNet-C   & 4 & 61 & 0.7 &  2.05 	\\
	DenseNet-C   & 4 & 61 & 0.8 &  2.02 	\\
    DenseNet-C	 & 4 & 61 & 0.9 &  2.16 	\\
	\noalign{\hrule height 1pt}
	\end{tabular}
    \caption{Comparison of networks with different compression parameter ($\theta$) in terms of WER (\%) on RM test set and the training time when systems were trained with 3 hours RM data set.}
    \label{table:rm_densenet_compression}
\end{table}

\begin{table}[t]
\scriptsize
    \centering
	\begin{tabular}{llcccl}
    \noalign{\hrule height 1pt}
	\text{System} & \text{\# of} & \text{depth} & \text{$\theta$} & \text{WER}   & \text{Train}   \\
    	   & \text{db.}  &	& &	  & \text{time}			\\
    \hline
           &   &&&\textit{dev'93 eval'92}&\text{(days)}\\
    \hline
	DenseNet	  & 3 & 22 & 1.0  &  9.51 5.41 & 	7	\\
	DenseNet-C   & 3 & 22 & 0.5 &  9.76 5.33 &	6	\\
    DenseNet-BC   & 3 & 22 & 0.5 &  12.04 6.91 &	3	\\
    \hline
	DenseNet      & 3 & 41 & 1.0 &  7.69 4.41 &		17\\
	DenseNet-C   & 3 & 41 & 0.5 &  7.78 4.25 &	10	\\
    DenseNet-BC   & 3 & 41 & 0.5 & 9.39 5.23  &	5	\\
    \hline
    DenseNet-C   & 4 & 41 & 0.5 &  7.74 4.22  & 5 \\
	DenseNet-C   & 4 & 61 & 0.4 &  \textbf{7.12 3.81} & \textbf{10} \\
    \noalign{\hrule height 1pt}
	\end{tabular}
    \caption{Comparison of networks with different hyper-parameters in terms of WER (\%) on \textit{dev'93} and \textit{eval'92} and the training time when systems were trained with 78 hours WSJ data set.}
    \label{table:wsj}
\end{table}

\begin{table}[t]
\small
    \centering
	\begin{tabular}{llccc}
    \noalign{\hrule height 1pt}
	\text{data set} & \text{System} & \text{Depth} & \text{Param.}& \text{WER}  \\
	\hline
    &DNN			 & 6  & 8M & 2.41\\
    RM&CNN			 & 7  & 9M & 2.36\\
     &VGG			 & 14 & 20M& 2.29\\
    &DenseNet-C   	 & 61 & 1M& \textbf{1.91} \\
	\hline
    \hline
    &&&&\textit{\small dev'93 eval'92}\\
    \hline
    &DNN			 & 6  & 8M& 8.61 4.82\\
    WSJ&CNN			 & 7  & 9M& 8.49 4.57\\
    &VGG		     & 14 & 20M& 8.80 5.19\\
    &DenseNet-C   	 & 61 & 1M& \textbf{7.12 3.81}\\
	\noalign{\hrule height 1pt}
	\end{tabular}
    \caption{The summary of baseline systems and DenseNet-C trained with 3 hours RM data set or 78 hours WSJ data set.}
    \label{table:summary}
\end{table}

\subsection{DenseNet}
Table ~\ref{table:DenseNet_architecture} presents the architectures of DenseNet-BC, Dense-Net and DenseNet-C with depth 22. The first layer of three networks is $3 \times 3$ convolution which is followed by 3 dense blocks. For DenseNet-BC, each dense block consists of multiple pairs of $1 \times 1$ convolution as bottleneck layer and $3 \times 3$ convolution. On the other hand, the dense block in DenseNet and DenseNet-C only consists of multiple $3 \times 3$ convolution. For all of three networks, each dense block except the last one is followed by a transition which consists of $1 \times 1$ convolution and $2 \times 2$ average pooling. 

The size of input features is $3 \times 11 \times 40$ since the context window is \{5,1,5\} and the feature is 40-bins log-Mel filterbank with the first and the second time derivatives. The number of feature-maps for each layer is increased by growth rate $k$. For example, if we have growth rate $k=12$ and 3 layers in the dense block, then the numbers of feature-maps of the first, second and third layers are 12, 24, 36, respectively. The output size of the first convolution layer with kernel size (3,3) and stride=1 is $9 \times 38$. The input and output size of dense block are the same because it will concatenate the input network and the output network of $3 \times 3$ convolution in dense block along the axis which is denoted as the number of feature-maps. Every bottleneck layer produces $4k$ feature-maps and we set $k=12$ as in ~\cite{densely-connected-convolutional-networks}. If a dense block contains $c$ feature-maps,  we let the following transition layer generate $\theta c$ output feature-maps. Note that DenseNet-BC and DenseNet-C have $\theta < 1$ and DenseNet has $\theta=1$. In this experiment, we set $\theta=0.5$ when investigating the effect of compression component of DenseNet. Afterwards, we set $\theta$ from $0.1$ to $0.9$ when trying to find the optimal performance for DenseNet-C.

The number of convolution layers in each Dense block is depending on the depth and the number of dense blocks. For example, if a DenseNet or DenseNet-C has depth 22 with 3 dense blocks, then there are $\dfrac{(22-4)}{3}=6$ $3 \times 3$ convolution layers in a dense block. Likewise, there are $\dfrac{6}{2}=3$ pairs of bottleneck layer and $3 \times 3$ convolution layer in a dense block. Both are mentioned in Table ~\ref{table:DenseNet_architecture}. The learning rate decaying schedule that starts with an initial learning rate of 0.01. All models were trained for around 20 epochs using learning rate decaying schedule. 

\section{Results}
Table ~\ref{table:rm_1} shows the comparison of the performance of Dense-Net, DenseNet-C and DenseNet-BC with 3 hours RM data set. We fixed the number of dense blocks (\# of db.) to 3 which has the best performance in Computer Vision~\cite{densely-connected-convolutional-networks} and the depth to 22. As expected, DenseNet-C and DenseN-et-BC were faster than DenseNet for training because they reduced the number of feature-maps. BenseNet-C perform-ed the best and DenseNet-BC got the worst WER. This is also happened in the original paper, DenseNet-BC performed worse than DenseNet when they have the same depth and growth rate $k$ ~\cite{densely-connected-convolutional-networks}. Then we increased the depth to 41, all of the three systems performed slightly better than the system with depth 22. However, the training time were also increased. The reason is that the number of feature-maps increases when the depth increases, and this number is one of factors affecting the number of trainable parameters. To improve the training time, we increased the number of dense blocks, in other words, increased the number of transition layers which do compression. From Table~\ref{table:rm_2}, DenseNet-C with depth 41 and 4 dense blocks has WER 2.10 which is similar to the best WER 2.12 reached by DenseNet-C with depth 41 and 3 dense blocks. Moreover, the training time is highly improved. There are two reasons why it took much less training time. Firstly, DenseNet-C with 4 dense blocks do more times of compressions which makes lesser number of parameters to train. Table~\ref{table:param} shows that DenseNet-C with 4 dense blocks has less number of parameters than the number from DenseNet-C with 3 dense blocks. Secondly, it took 12 Epochs to train DenseNet-C with 4 dense blocks and 16 Epochs to train DenseNet-C with 3 dense blocks. It seems that DenseNet-C with 4 dense blocks reaches the optimal faster than the system with 3 dense blocks. Knowing the fact that increasing the depth can improve the system performance shown in Table~\ref{table:rm_1}, we set the depth to 61, 81 and 101 for DenseNet-C with 4 dense blocks. Table~\ref{table:rm_2} shows that the network with depth=61 is optimal setting considering the WER and training time. 

DenseNet-C performs better than DenseNet and Denst-Net-BC with RM data set, but this system uses $\theta=0.5$ which we do not know whether it is optimal or not. Table~\ref{table:rm_densenet_compression} shows the comparison of systems performance with $\theta$ from $0.1$ to $0.9$. DenseNet-C with $\theta=0.4$ reaches the best WER 1.91. 

Table~\ref{table:wsj} shows the comparision of the performance of DenseNet, DenseNet-C and DenseNet-BC with 78 hours Wall Street Journal data set. In the previous experiments on RM data set, we also fixed the depth to 22 or 41 and the dense blocks to 3. The DenseNet-C with depth 41 performed the best with 10 days training time. After increasing the number of dense blocks, the training time got improved a lot without losing much performance. Then we tried DenseNet-C with depth 61 and $\theta=0.4$ which reached the best WER with the RM data set, the result shows that this model still performed well on WSJ data set. 

The comparison of baseline models (DNN, CNN and VGG) and DenseNet-C is shown in Table ~\ref{table:summary}. All of the four systems used the speaker independent FBANK features. 
DenseNet-C outperformed all other neural models on RM and WSJ data sets by a large margin. 

\begin{figure}[t]
  \centerline{\includegraphics[width=1.05\columnwidth]{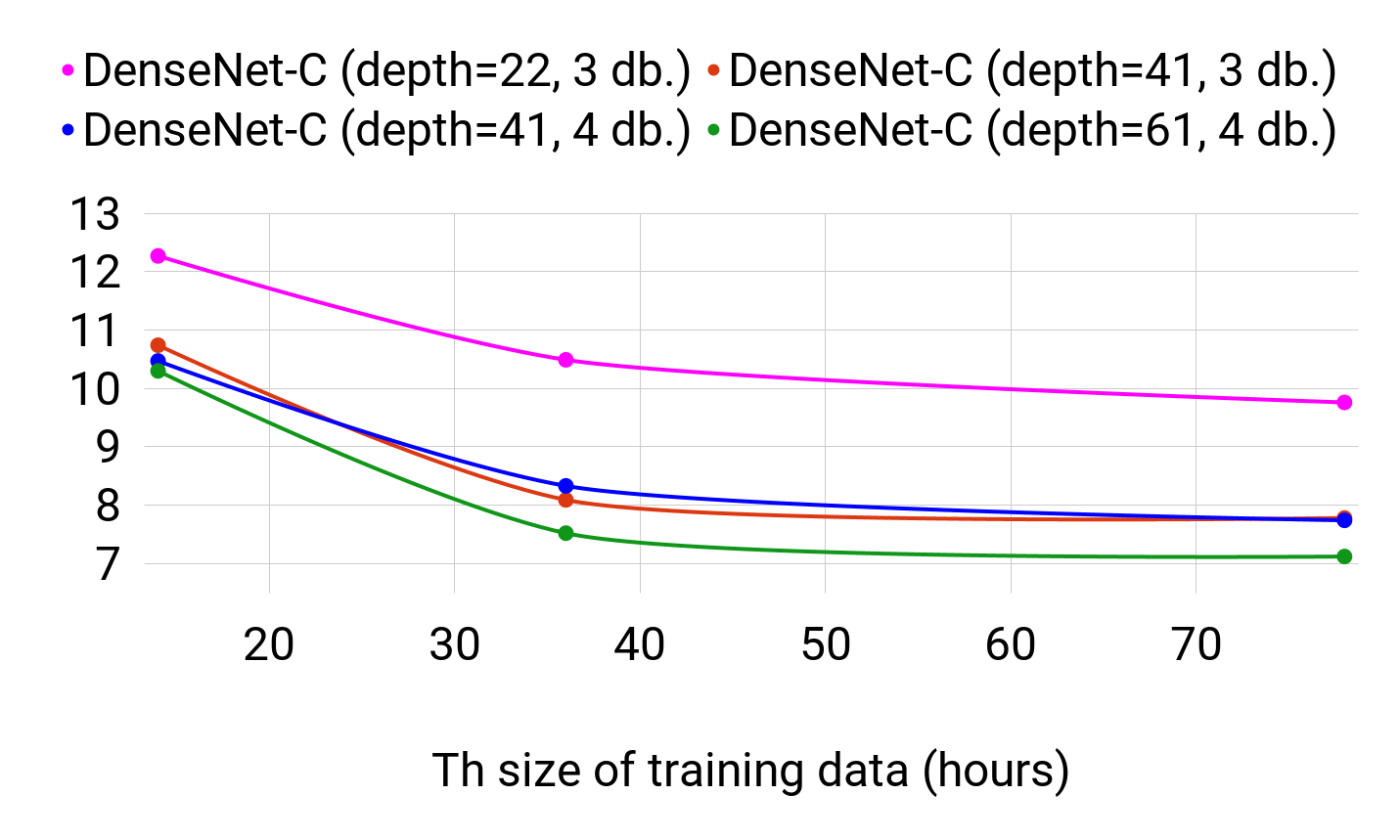}}
  \caption{The systems performance with different size of WSJ data. The vertical axis represents WER(\%) on $dev'93$.}
  \label{fig:densenet_trend}
\end{figure}

To explore the impact of training size on the final performance, we trained DenseNet-C with 14 hours (train\_si8-4, 83 speaker, 7183 utterances), 36 hours (283 speakers, 18666 utterances) and 78 hours (train\_si284: 283 speakers, 37416).
Fig~\ref{fig:densenet_trend} shows that the WER (\textit{dev'93}) 7.52 of DenseNet-C with depth=61 trained with 36 hours (green line) is close to the optimal WER (\textit{dev'93}) 7.12 of the same system trained with 78 hours. 
It implies that DenseNet with appropriate settings can almost achieve the optimal result without training on the entire WSJ data set.
Note that this WER is much better than the performance of all other neural models trained with the full data set.

\section{Conclusion}
This paper has successfully applied DenseNet for speech recognition. 
We explored the impact of different network design choices on the final ASR performance.
We found that the depth plays an important role and the bottleneck and compression components are helpful to shorten the training time.
Furthermore, results on two data sets (RM and WSJ) showed that DenseNet significantly outperforme-d other neural models such as DNNs, CNNs and VGGs.
Finally, due to compact architecture with small amount of trainable parameters, we were able to train DenseNet with only a half of the WSJ data set outperforming other models trained with the full data set by a large margin. We will compare DenseNets with state-of-the-art models like BLSTM and TDNN using unadapted feature in the future.

\small
\newpage
{}


\end{document}